%% file: paper_template.tex
\documentclass[conference]{IEEEtran}
\IEEEoverridecommandlockouts
\usepackage{times}
\usepackage[numbers]{natbib}
\usepackage{multicol}
\usepackage[bookmarks=true]{hyperref}
\usepackage{graphics} % for pdf, bitmapped graphics files
\usepackage{epsfig} % for postscript graphics files
\usepackage{times} % assumes new font selection scheme installed
\usepackage{amsmath} % assumes amsmath package installed
\usepackage{xcolor}

\usepackage{mathrsfs}
\usepackage{amssymb}
\usepackage{dsfont}
\usepackage{xcolor}
\usepackage{mathtools}
\usepackage{mathrsfs}
\usepackage{subfig}
\usepackage{balance}
\usepackage{booktabs}
\usepackage{multirow}
\usepackage[ruled]{algorithm}% http://ctan.org/pkg/algorithms
\usepackage[noend]{algpseudocode}
\newcommand{\mathbbm}[1]{\text{\usefont{U}{bbm}{m}{n}#1}}
\algnewcommand\algorithmicinput{\textbf{Input:}}
\algnewcommand\INPUT{\item[\algorithmicinput]}

\algnewcommand\algorithmicoutput{\textbf{Output:}}
\algnewcommand\OUTPUT{\item[\algorithmicoutput]}

\algnewcommand\algorithmicinitialize{\textbf{Initialize:}}
\algnewcommand\Initialize{\item[\algorithmicinitialize]}

\newtheorem{remark}{Remark}

\newcommand{\ctrlseq}{\mathbf{u}}

\begin{document}

\title{Bridging Model Predictive Control and Deep Learning for Scalable Reachability Analysis}

\author{Zeyuan Feng$^{1}$, Le Qiu$^2$, and Somil Bansal$^{1}$% 
\thanks{$^{1}$Authors are with the Department of Aeronautics and Astronautics at Stanford University, USA {:\{zeyuanf, somil\}@stanford.edu}.}
\thanks{$^{2}$Author is with the Department of Electrical Engineering at Tsinghua University, China:
        {\{qiule1026@gmail.com\}}.}%
\thanks{This research is supported in part by the DARPA Assured Neuro Symbolic Learning and Reasoning (ANSR) program and by the NSF CAREER program (2240163).}%
\thanks{The implementation can be found on \url{https://github.com/smlbansal/deepreach/tree/DeepReach_MPC}.}
}
\maketitle

\begin{abstract}
Hamilton-Jacobi (HJ) reachability analysis is a widely used method for ensuring the safety of robotic systems. 
Traditional approaches compute reachable sets by numerically solving an HJ Partial Differential Equation (PDE) over a grid, which is computationally prohibitive due to the \textit{curse of dimensionality}.
Recent learning-based methods have sought to address this challenge by approximating reachability solutions using neural networks trained with PDE residual error. 
However, these approaches often suffer from unstable training dynamics and suboptimal solutions due to the weak learning signal provided by the residual loss.
In this work, we propose a novel approach that leverages model predictive control (MPC) techniques to guide and accelerate the reachability learning process. 
Observing that HJ reachability is inherently rooted in optimal control, we utilize MPC to generate approximate reachability solutions at key collocation points, which are then used to tactically guide the neural network training by ensuring compliance with these approximations.
Moreover, we iteratively refine the MPC-generated solutions using the learned reachability solution, mitigating convergence to local optima. 
Case studies on a 2D vertical drone, a 13D quadrotor, a 7D F1Tenth car, and a 40D publisher-subscriber system demonstrate that bridging MPC with deep learning yields significant improvements in the robustness and accuracy of reachable sets, as well as corresponding safety assurances, compared to existing methods.
\end{abstract}

\IEEEpeerreviewmaketitle

\sloppy                % Loosens line breaking
\interlinepenalty=100  % Allow page breaks within paragraphs
\clubpenalty=100       % Allow breaks after first line of paragraph
\widowpenalty=100      % Allow breaks before last line of paragraph
\displaywidowpenalty=100

\section{Introduction}
\label{sec:intro}
\input{intro}

\section{Preliminaries}
\label{sec:preliminaries}
\input{preliminaries}

\section{Learning Reachability Solutions with MPC-Based Guidance}
\label{sec:approach}
\input{approach}

\section{Experiments}
\label{sec:results}
\input{results}

\section{Conclusion}
\label{sec:conclusion}
\input{conclusion}

\bibliographystyle{plainnat}
\bibliography{references,reachability}

\section{Appendix}
\label{sec:appendix}
\input{appen}

\end{document}

%% file: intro.tex
Hamilton-Jacobi (HJ) reachability analysis is one of the most widely used tools for providing formal safety assurances for autonomous systems. 
It characterizes the unsafe states of the system via Backward Reachable Tube (BRT) -- the set of all initial states from which a system failure is inevitable.
Thus, the complement of the BRT provides a safe set for the system. 
Along with the safe set, reachability analysis provides a safety controller for the system that can be used as it is or alongside a nominal (potentially data-driven) controller to maintain safety.

In HJ reachability, the BRT computation is framed as an optimal control problem that results in solving a certain Hamilton-Jacobi-Bellman PDE (HJB-PDE) \cite{mitchell2005time,Margellos11}.
Solving this PDE yields a safety value function that implicitly represents both the BRT and the safety controller.
Consequently, a number of methods have been developed to solve the HJB PDE.
Traditional methods solve HJB PDE numerically over a grid, which suffers from the so-called ``curse of dimensionality'' \cite{bansal2017hamilton}.
Specifically, the computation scales exponentially with the system dimension, making systems of more than 6D intractable.
Many techniques for speeding up reachability analysis put restrictions on the type of system allowed and/or assign predefined shapes to the safe set (e.g. ellipsoids, polytopes) \cite{Darbon2016HopfLax, Chow2017HopfLax, Frehse2011SpaceEx, Kurzhanski2002Ellipsoidal, Maidens2013Lagrangian, Majumdar2013SOS, Dreossi2016Parallelotope, Henrion2013Polynomial}. However, computing reachable sets for general nonlinear systems remains a challenge.
This motivated the recent development of learning-based methods to approximate the HJB-PDE solution \cite{nakamura2021adaptive}. 
One line of research leverages Reinforcement Learning (RL) to approximate the safety value function, achieving impressive performance improvements \cite{fisac2019bridging,hsu2021safety,hsu2023isaacs,li2025certifiabledeeplearningreachability}.
Another line of work \cite{9561949,singh2024imposing}, which this paper aims to enhance, learns value function via self-supervised learning by minimizing the residuals of HJB-PDE. 
The latter set of approaches, termed DeepReach variants, are rooted in recent advances in Physics-Informed Machine Learning \cite{RAISSI2019686, li2022physicsinformed} and have the theoretical advantage of recovering the exact HJB-PDE solution as the training loss converges to zero \cite{hofgard2024convergence}.
However, in practice, these methods often converge to suboptimal solutions due to the weak learning signal provided by the residual loss alone. 
Additionally, as we will demonstrate, this can result in non-physical solutions, where the learned value function significantly deviates from the ground truth, even when training loss appears low. 
Consequently, these methods exhibit instability and inaccuracies, especially for stiff dynamical systems or problems with complex boundary conditions (i.e., intricate safety specifications). 
To combat these challenges, \cite{sharpless2024linear} leverages the Hopf formula to generate HJB-VI solutions for linearized dynamics and learns the solution using semi-supervision. However, this method does not synthesize value function labels for the original nonlinear dynamics and relies heavily on the quality of the Hopf solutions. 
Another method imposes exact safety specification  \cite{singh2024imposing}, but it offers limited improvements, and learning instability for general nonlinear systems remains a key challenge.

\begin{figure*}[t!]
		\centering
         {\includegraphics[width=0.8\textwidth]{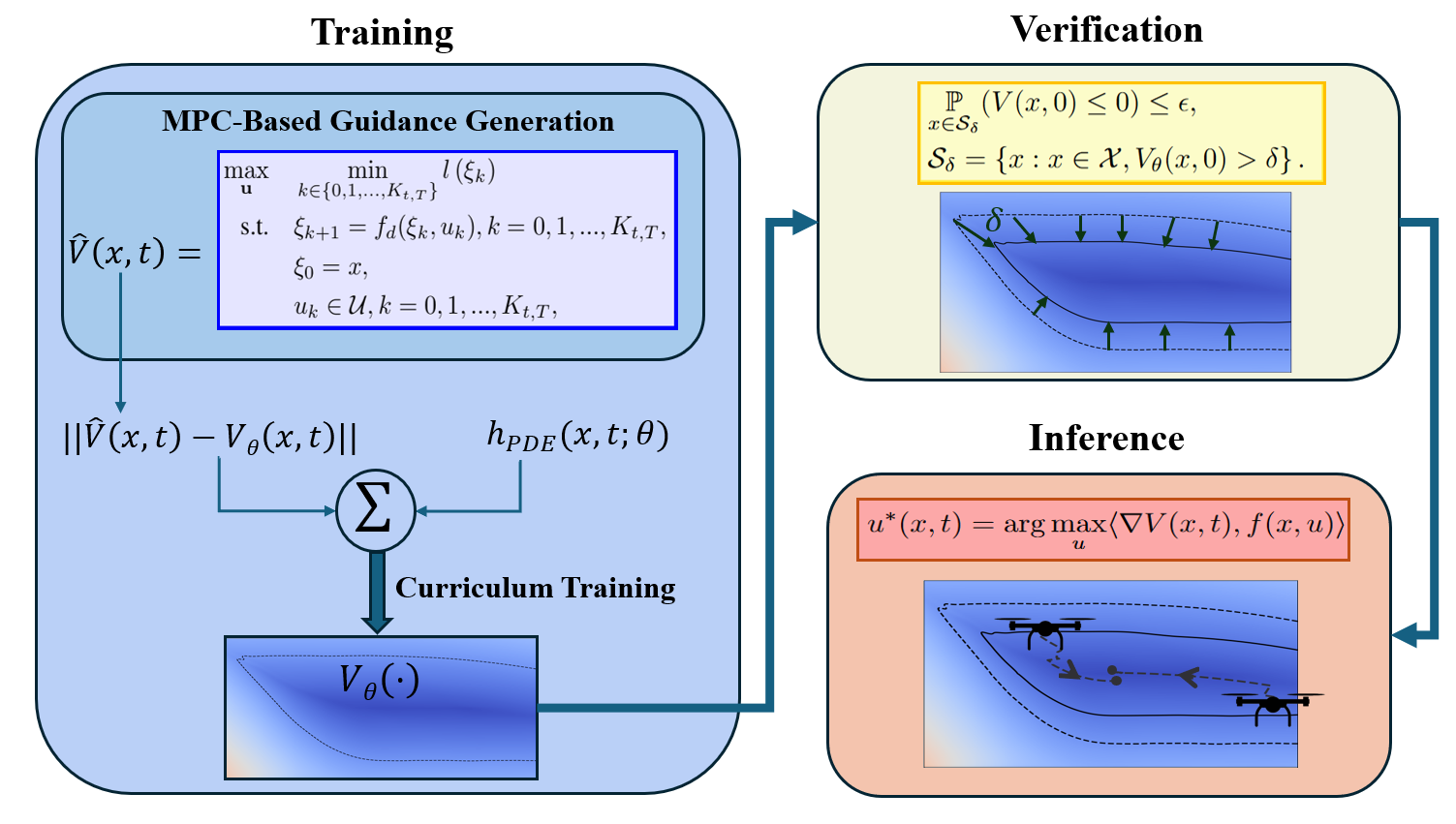}}
 	\caption{\small{We propose a framework that efficiently generates approximate safety value function datasets using a sampling-based MPC approach and integrates these data labels to guide the learning of reachability solutions for high-dimensional autonomous systems. The learned value function is then verified using conformal prediction, providing probabilistic safety assurances for the system under the induced safe policy.}}
    \vspace{-1.5em}
 	\label{fig:front_fig}
\end{figure*}

To overcome these challenges, we propose a framework that leverages approximate reachability solutions to guide the learning of the safety value function.
Our approach is inspired by the success of incorporating data-driven loss terms in Physics-Informed Neural Networks (PINNs) \cite{cai2021physics}, which effectively anchors trial solutions at key collocation points alongside PDE residuals, thereby accelerating convergence and guiding learning toward feasible solutions.
Specifically, building on the optimal control foundations of HJ reachability, we propose a sampling-based MPC approach to generate approximate solutions for high-dimensional reachability problems.
These solutions serve as semi-supervised labels to improve value function learning.  
Our method is designed to be highly parallelizable on GPUs, allowing for efficient synthesis of large datasets. 
Moreover, we provide an algorithm to iteratively update the MPC dataset during training using the learned DeepReach policy, progressively improving accuracy and stability of the learned value function.
By combining sampling-based MPC with DeepReach-style self-supervised learning, this work integrates the strengths of both approaches -- leveraging MPC for efficient and structured dataset generation and DeepReach for reliable safety assurances.
To summarize, our overall contributions are:
\begin{itemize}
    \item A novel data-generation approach that efficiently computes approximate HJB-VI solutions using an MPC-based sampling method.
    \item A hybrid training algorithm that balances residual loss with data-driven supervision to improve the convergence of value function learning, while iteratively refining the MPC dataset during training.
    \item Demonstration of the proposed approach on four challenging reachability problems: a 2D vertical drone avoiding ground and ceiling, a 13D quadrotor avoiding a cylindrical obstacle, a 7D F1Tenth car navigating a track, and a 40D publisher-subscriber system, highlighting a significant improvement in accuracy and stability of value function learning. 
\end{itemize}

%% file: preliminaries.tex
In this section, we formulate the reachability problem and provide a brief background on HJ reachability. 

\subsection{Problem Setup}
We consider a dynamical system with time-invariant, possibly nonlinear, dynamics: $\dot{x}=f(x,u)$,
where $x \in \mathbb{R}^n$ is the state and $u \in \mathcal{U} \subset \mathbb{R}^{n_u}$ is the control input.
The safety requirement is defined by a Lipschitz-continuous function $l: \mathbb{R}^n \rightarrow \mathbb{R}$, with its sub-zero level set being the failure set, i.e., $\mathcal{L} = \{x:  l(x) \leq 0\}$.
For example, in a robotic navigation setting, $\mathcal{L}$ may represent obstacles, while $l(x)$ could be the signed distance function to these obstacles.

Our primary goal in this work is to compute the BRT of the system, denoted as $\mathcal{B}$, which consists of all initial states from which the system will inevitably enter the failure set $\mathcal{L}$ within the time horizon $[0, T]$, regardless of the control strategy. Formally:
\begin{equation}
\label{eq:avoid_BRT}
    \mathcal{B}=\left\{x: \forall u(\cdot) \in \mathcal{U}, \exists \tau \in[0, T], \xi_{x,0}^{u(\cdot)}(\tau) \in \mathcal{L} \right\},
\end{equation}
where $\xi_{x,t}^{u(\cdot)}(\tau)$ is the state achieved at time $\tau$ by applying the control policy $u(\cdot)$, starting from initial state $x$ at time $t$.
Thus, $\mathcal{B}$ captures all states from which the system is doomed to fail. The complement of the BRT,
$\mathcal{S} := \mathcal{B}^C$, represents the safe set for the system, i.e., the initial states from which it is possible to avoid entering the failure region. 
Correspondingly, our second objective is to synthesize a safe control policy $u^*(\cdot)$ that ensures the system remains in the safe set $\mathcal{S}$ and does not enter $\mathcal{B}$, whenever possible.

\subsection{Computation of BRT Using HJ Reachability}
In HJ reachability, the computation of $\mathcal{B}$ is formulated as a continuous-time optimal control problem, with the following cost function:
\begin{equation} \label{eqn:safety_cost}
    J(x, t, u(\cdot)) =  \min _{\tau \in[t, T]} l\left(\xi_{x,t}^{u(\cdot)}(\tau)\right).
\end{equation}
Here, $J(x, t, u(\cdot))$ represents the minimum ``distance'' to the failure set along the state trajectory starting at $x$ and governed by $u(\cdot)$ over $[t, T]$.
The value function and the optimal controller can be obtained by maximizing this cost function:
\begin{equation}
    \begin{aligned}
        V(x, t) &=  \sup _{u(\cdot) \in \mathcal{U}_{[t, T]}} J(x, t, u(\cdot)), \\
        u^*(\cdot) & = \arg \sup _{u(\cdot) \in \mathcal{U}_{[t, T]}} J(x, t, u(\cdot)).
    \end{aligned}
\end{equation}

Once $V(x, t)$ is computed, the BRT is given by the sub-zero level set of the value function:
\begin{equation}
\label{eq:brt_from_V}
\mathcal{B}=\left\{x: V(x,0)  \leq 0 \right\}.
\end{equation}
Mathematically, BRT consists of all states where the minimum distance to the failure set is negative under the optimal control. 
In other words, the system must have entered the failure set at some time in $[t, T]$ and hence these states are contained in the BRT.

The value function $V(x, t)$ can be computed using the principle of dynamic programming which results in the following Hamilton-Jacobi-Bellman Variational Inequality (HJB-VI):
\begin{equation}
\label{eq:HJB-VI}
\begin{gathered}
     \min \{D_{t}V(x,t) + H(x,t), l(x) - V(x,t) \} = 0, \\
     V(x,T) = l(x),\\
     H(x,t) = \max_{u \in \mathcal{U}} \langle \nabla V(x,t) \; , \; f(x,u) \rangle,
\end{gathered}
\end{equation}
where the second line specifies the boundary condition, $H(x,t)$ is the Hamiltonian, $D_{t}$ represents the time derivatives and $\nabla$ denotes the spatial derivatives of the value function.
For more details on HJ reachability analysis and the derivation of HJB-VI, we refer readers to \cite{lygeros2004reachability,mitchell2004toolbox, bansal2017hamilton}.

With the value function in hand, the optimal safety controller to keep the system outside the BRT is given by:
\begin{equation}\label{eq:safe_control}
    u^*(x,t)=\arg \max _u \langle\nabla V(x, t), f(x, u)\rangle.
\end{equation}

For low-dimensional systems, HJB-VI can be solved numerically on a state-space grid using various toolboxes \cite{mitchell2004toolbox, mitchell2005time, lygeros2004reachability, Margellos11,fisac2015reach}. However, for high-dimensional systems, these methods suffer from the curse of dimensionality; consequently, learning-based methods have been proposed to solve HJB-VI. This work builds upon one such line of learning methods called DeepReach \cite{9561949,singh2024imposing,chilakamarri2024reachability}, that learn the HJB-VI solution for high-dimensional problems in a self-supervised manner by minimizing the following HJB-VI residuals:
\begin{equation} \label{deepreach_loss}
\begin{aligned}
 \mathop{\mathbb{E}}_{(x_i,t_i) \in \mathcal{X} \times [0,T]} &\left[  \| \min \left\{D_t V_\theta\left(x_i, t_i\right)  +H\left(x_i, t_i\right), \right. \right.  \\
& \qquad \qquad \left. \left. l\left(x_i\right)-V_\theta\left(x_i, t_i\right)\right\} \| \right].
\end{aligned}
\end{equation}
Here, the value function is approximated as $V_\theta(x,t)=l(x)+(T-t)\cdot O_\theta(x,t)$, where $O_\theta(x,t)$ is the output of a NN with trainable parameters $\theta$. This formulation inherently satisfies the boundary condition $V_\theta(x,T) = l(x)$, thereby leaving only the PDE residual errors in the loss function (\ref{deepreach_loss}). Again, we emphasize that learning the value function for complex, high-dimensional reachability problems using pure self-supervision remains challenging and unstable. To address this, we aim to leverage approximated value function data to enhance the training process.

%% file: approach.tex
Our approach consists of three key steps: first, we generate an approximate safety value function using a sampling-based MPC method that optimizes the safety cost function in \eqref{eqn:safety_cost}.
These approximate value samples are then used to augment the training loss of DeepReach to incentivize the learned value function to be consistent with them.
Finally, the MPC-guided value function is corrected for potential learning errors using a conformal prediction scheme, thereby providing a verified safe set of the system.
We now explain each of these steps in detail.
An overview of our approach is in Fig.~\ref{fig:front_fig}.

\subsection{Generating Approximate Value Function Dataset Using Sampling-Based MPC}
Our key idea is that HJ reachability computes the BRT via formulating it as an optimal control problem where the cost function is given by \eqref{eqn:safety_cost}.
Thus, an approximate safety value function can be obtained by solving this optimal control problem using alternative optimal control methods, such as MPC. Subsequently, this approximate value function can be leveraged to enhance the training of DeepReach.

Specifically, we compute an approximate value function, $\hat{V}$, by solving the following discrete-time version of the optimal control problem:
\begin{equation}
\label{eq:mpc_prob}
\begin{aligned}
\hat{V}(x, t) = \max_{\ctrlseq} \quad & \min _{h \in \left\{0, 1,..., H_{t,T} \right\}} l\left(\xi_{h}\right)\\
\textrm{s.t.} \quad & \xi_{h+1}= f_d(\xi_h, u_h), h= 0,1,..., H_{t,T},\\
  & \xi_0 = x, \\
  & u_h \in \mathcal{U}, h= 0,1,..., H_{t,T},
\end{aligned}
\end{equation}
where $h \in \left\{0, 1,..., H \right\}$ denotes the time steps between $t$ and $T$, $\xi_{h}$ is the system state, and $\ctrlseq := [u_0, \cdots, u_H]$ is the control sequence. 
$f_d$ are the discretized dynamics that can be obtained from the continuous dynamics using first-order Euler approximation.
The optimization problem in \eqref{eq:mpc_prob} can be solved using a broad class of MPC methods. 
Specifically, we leverage sampling-based MPC methods to solve this discrete-time problem, as they are simple, fast, and highly parallelizable.

Our MPC algorithm is summarized in Alg.~\ref{alg:MPC_data}.
The algorithm takes as inputs the number of initial states $|D_{MPC}|$, the number of hallucinated trajectories $N$, the number of iterative sampling steps $R$, the time horizon $H$, the nominal control $\ctrlseq_{nom}$, the time step $\Delta$, and the state space of interest $\mathcal{X}$.
The output of the algorithm is a dataset $\mathcal{D}_{MPC}$, where each data point consists of an initial time $t$, an initial state $x$, and a corresponding value function approximation $\hat{V}(t, x)$.
To generate a single data point, the algorithm begins by sampling $N$ distinct control sequences, $\tilde{\ctrlseq}_n$, around the nominal control sequence $\ctrlseq_{nom}$. 
These $N$ trajectories are rolled out using discretized dynamics with a time step of $\Delta$, and the corresponding cost values $J_n$ are computed. 
Finally, the nominal control sequence is updated to the best-performing control sequence, and the process is repeated for $R$ iterations. 
This process is highly parallelizable in practice and can be significantly accelerated with modern GPU computation.
\begin{algorithm}[H]
\caption{MPC Dataset Generation}\label{alg:MPC_data}
\begin{algorithmic}[1]
\INPUT $|D_{MPC}|$, $N$, $R$, $H$, $\mathbf{u}_{nom}$, $\Delta$, $\mathcal{X}$
\OUTPUT $\mathcal{D}_{MPC} = \left\{ (t_i, x_i, \hat{V}(t_i, x_i))\right\}$%$\mathcal{D}_{MPC} = \left\{ (t_i, x_i, \hat{V}(t_i, x_i))\right\}_{i=1}^{A}$
\Initialize $\mathcal{D}_{MPC}=\emptyset$
\For{$i=1:|D_{MPC}|$}
    \State $x_i \sim \text{Uniform}(\mathcal{X})$; $t_i \sim \text{Uniform}(0, T)$
    \For{$r=1:R$}
        \For{$n=1:N$}
            \State $\tilde{\mathbf{u}}^n \sim \text{Gaussian}(\mathbf{u}_{nom}, \sigma^2)$
            \State $\xi^n_0 \leftarrow x_i$
            \For{$h=0:H-1$}
                \State $\xi^n_{h+1}=\xi^n_{h}+f(\xi^n_{h},\tilde{\mathbf{u}}^n_h)\Delta$
            \EndFor
            \State  $J^n= \min_{h=0,1,...,H} l(\xi^n_{h})$ 
        \EndFor
        \State $n^*=\arg\max_n J^n $ 
        \State $\hat{V}(t_i, x_i) \leftarrow  J^{n^*} $ 
        \State $\mathbf{u}_{nom} \leftarrow \tilde{\mathbf{u}}^{n^*}$
    \EndFor
    \State $\mathcal{D}_{MPC} \leftarrow \mathcal{D}_{MPC} \cup (t_i, x_i, \hat{V}(t_i, x_i))$
\EndFor
\end{algorithmic}
\end{algorithm}

We conclude this section with a few remarks about the proposed MPC method.
\begin{remark}
    In practice, we set one of the sampled control sequences in Line-5 to $\mathbf{u}_{nom}$. This allows for a monotonic improvement in the MPC value function over $R$ iterations. 
\end{remark}

\begin{remark}
    We can efficiently bootstrap the MPC dataset by computing the running value function along the optimal state-control trajectory, terminating when the minimum $l(x)$ is reached. 
    Specifically, if $\xi^{n^*}$ denotes the state trajectory under the optimal control sequence $\tilde{\mathbf{u}}^{n^*}$, we can estimate the value function at intermediate points along this trajectory as:
    \begin{equation*}
        \hat{V}(\Delta \cdot h,\xi^{n^*}_h)= \min_{j = h,\ldots,H} l(\xi^{n^*}_{j}).
    \end{equation*}
    Accordingly, we add the data point $(\Delta \cdot h, \xi^{n^*}_h, \hat{V}(\Delta \cdot h,\xi^{n^*}_h))$ to $\mathcal{D}_{MPC}$.
    This approach enables rapid collection of a large, high-quality MPC dataset for guiding the learning process.
\end{remark}

\begin{remark}
    Several design choices remain for the MPC algorithm, including how to compute the next control input (e.g., weighted sum vs. best sample) and the choice of control perturbation distribution (e.g., uniform vs. Gaussian). Empirically, we found that selecting the best sequence and using Gaussian sampling yielded the least noisy datasets. However, these options remain configurable, allowing users to tailor the approach to their specific needs.
\end{remark}

\subsection{Learning Reachability Solution Using MPC Dataset and HJB-VI Residuals}
Our approach (summarized in Algorithm~\ref{alg:training}) leverages a three-phase learning scheme for learning the safety value function: pretraining, curriculum training, and fine-tuning. 
The goal is to train a value function that accurately captures the BRT, using data-driven supervision from the MPC dataset and HJ reachability-based PDE residual loss.
\vspace{1em}

\noindent \textbf{Pretraining Phase: Supervised Learning from MPC Dataset.} 
During the pretraining phase, the model is trained using the collected MPC dataset, which consists of sampled states and their corresponding estimated value function. 
The training objective minimizes the prediction error between the learned value function $V_{\theta}(x,t)$ and the approximated value function:
\begin{equation} \label{eqn:data_loss}
\begin{aligned}
    &\mathcal{L}_{data}=\sum_{i=1}^{N_{MPC}} h_{data}(x^{MPC}_i,t^{MPC}_i,\hat{V_i}; \theta), \\
    &h_{data}(x,t,\hat{V}; \theta) = \|\hat{V}(x, t)- V_\theta(x,t) \|.
\end{aligned}
\end{equation}

This supervised learning step aligns the learned value function with the approximated reachability solution inferred from the MPC dataset, providing an informative ``warmstart'' of safe and unsafe regions for the neural network.
\vspace{1em}

\noindent \textbf{Curriculum Training: Joint Optimization with PDE Loss.}
After pretraining, we refine the value function using a curriculum training strategy that gradually increases the horizon of training samples, meaning the time horizon is progressively increased from $[T, T]$ to $[0, T]$ over $N_c$ iterations.  
Since HJB-VI is a terminal time PDE, the accuracy of safety value function at earlier time steps (corresponding to smaller $t$) depends on the accuracy of predictions at later times. 
Intuitively, this temporal curriculum reduces the complexity of the learning problem.
The loss function during curriculum training consists of two components: the data-driven loss ($\mathcal{L}_{data}$) from the MPC dataset and the HJB-VI residual loss ($\mathcal{L}_{PDE}$), which enforces the HJ reachability equation:
\begin{equation} \label{eqn:pde_loss}
    \begin{aligned}
    &\mathcal{L}_{PDE}=\sum_{i=1}^{N_{PDE}} h_{PDE}(x^{PDE}_i,t^{PDE}_i;\theta), \\
    &h_{PDE}(x,t;\theta)=\| \min \left\{D_t V_\theta\left(x, t\right) +  H\left(x, t\right), \right. \\
    & \qquad \qquad \qquad \qquad \qquad \left.  l\left(x\right)-V_\theta\left(x, t\right)\right\} \|.
    \end{aligned}
\end{equation}
The overall loss function during the curriculum learning phase is given by:
\begin{equation}
\mathcal{L}_{combined}=\mathcal{L}_{PDE}+\lambda \mathcal{L}_{data},
\end{equation}
where a trade-off factor $\lambda$ is dynamically adjusted based on the loss gradients to balance the contributions of $\mathcal{L}_{PDE}$ and $\mathcal{L}_{data}$, making them equally important.  

A crucial step in the curriculum training phase is MPC dataset refinement. The initial dataset generated by the sampling-based MPC method is inherently suboptimal and noisy for long horizons, which can hinder the convergence of the PDE loss and cause the model to overfit to outliers, leading to violations of the governing HJB-VI equations. To mitigate this, we periodically refine the MPC dataset using the learned value function.

Specifically, when the time curriculum reaches the effective horizon, $H_R$, of the current MPC dataset (i.e., when $t = t_R$), we leverage the learned value function and policy to generate a new MPC dataset with an extended time horizon of $[t_R - H_R, T]$.
This updated dataset is computed by incorporating the learned value function as the terminal cost in the MPC formulation, providing a more accurate estimate of the optimal cost-to-go over the horizon $[t_R, T]$ while keeping the effective MPC horizon fixed at $H_R$.
With this terminal cost, we repeat the MPC dataset generation procedure in Algorithm $\ref{alg:MPC_data}$, effectively obtaining the MPC value function approximation over the time horizon $[t_R - H_R, T]$.
This dataset is then used to continue the curriculum training for another $H_R$ seconds and the entire process is repeated again.
By iteratively refining the dataset every $H_R$ seconds, we ensure that the MPC dataset remains informative and consistent, continuously guiding the learning process with higher-quality supervisory signals.
\vspace{1em}

\begin{algorithm}[t!]
\caption{DeepReach Training with MPC Dataset}\label{alg:training}
\begin{algorithmic}[1]
\INPUT $N_{PDE}$, $N_{MPC}$, $\mathcal{D}_{MPC}$, $\alpha$, $N_c$, $H_R$ 
\OUTPUT $V_\theta(x, t)$
\Initialize $\lambda \leftarrow 1.0$, $t_R \leftarrow T-H_R$
\For{each pretraining step}
    \State $(t_i,x_i,\hat{V}_i) \sim Uniform(\mathcal{D}_{MPC})$, $i=1,2,...,N_{MPC}$
    \State Compute $\mathcal{L}_{data}$ using \eqref{eqn:data_loss} 
    \State $\theta \leftarrow \theta - \alpha \nabla_\theta \mathcal{L}_{data}$
    % Step the optimizer with $\mathcal{L}_{data}$
\EndFor
\For{$n_c=1:N_c$}
    \State $t=T \left(\frac{N_c-n_c}{N_c} \right)$ %\Comment{Current start time}
    \State $(x^{PDE}_i,t^{PDE}_i) \sim  \text{Uniform}( \mathcal{X} \times [t,T])$, $i=1:N_{PDE}$
    \State $(x^{MPC}_i,t^{MPC}_i,\hat{V}_i) \sim \text{Uniform}(\mathcal{D}_{MPC})$
    \State Compute $\mathcal{L}_{data}$ and $\mathcal{L}_{PDE}$ using \eqref{eqn:data_loss} and \eqref{eqn:pde_loss} 
    \State $\lambda \leftarrow 0.9\lambda + 0.1\frac{||\nabla_\theta \mathcal{L}_{PDE}||}{||\nabla_\theta \mathcal{L}_{data}||}$
    \State $\theta \leftarrow \theta - \alpha \nabla_\theta \mathcal{L}_{combined}$
    % \State Step the optimizer with $\mathcal{L}_{combined}=\mathcal{L}_{PDE}+\lambda \mathcal{L}_{data}$ 
    \If{$t<t_R$}
        % \State $\mathcal{D}_{MPC} \leftarrow DatasetGeneration( V_\theta(\cdot),\newline \hspace*{3.5em}  H=\lceil \frac{min(T-t_R+\H_R,T)}{\Delta} \rceil, \H_R=\lceil \frac{min(\H_R,t_R)}{\Delta} \rceil$)
        \State Refine $\mathcal{D}_{MPC}$ using  $V_\theta$
        \State $t_R \leftarrow t_R - H_R$
    \EndIf
\EndFor
\For{each fine-tuning step}
    \State Repeat the curriculum step with $t=0$ and \newline \hspace*{1.4em} $\mathcal{L}_{data} := \mathcal{L}_{data,FT}$ 
\EndFor
\end{algorithmic}
\end{algorithm}
\vspace{-.55em}

\noindent \textbf{Fine-Tuning: Reducing False Positives for Safety Guarantees.}
The final fine-tuning phase improves the accuracy of the learned solution by employing a smaller learning rate and refining the safety decision boundary. 
A key challenge is minimizing false positives, where the learned value function incorrectly predicts a state as safe but is actually driven to failure under the safe policy induced by the value function.
Such errors can lead to catastrophic consequences in real-world safety-critical systems. 

To address this, the fine-tuning process introduces a modified data-driven loss function, where the loss for false positives is amplified by a scaler parameter $\lambda_{FP}$. 
This ensures that safety predictions remain conservative, reducing the likelihood of unsafe behaviors.
The loss function is defined as follows:
\begin{equation} \label{eqn:finetune_loss}
\begin{aligned}
   &\mathcal{L}_{data,FT}=\sum_{i=1}^{N_{MPC}} h_{FT}(x^{MPC}_i, t^{MPC}_i,\hat{V};\theta), \\
    &h_{FT}(x, t, \hat{V};\theta) \\
    & \ \ = 
        \begin{cases}
            \lambda_{FP} \|\hat{V}- V_\theta(x,t) \|  V_\theta(x,t), & \text{if } V_\theta(x,t) \geq 0 \land \hat{V}<0 \\
            \|\hat{V}- V_\theta(x,t) \|.              & \text{otherwise}
        \end{cases} 
\end{aligned}
\end{equation}
Note that since fine-tuning is done after the curriculum learning phase, MPC is able to use the learned safety policy (induced by the learned value function via Eqn~(\ref{eq:safe_control})) to generate an informative nominal control sequence for computing $\hat{V}$. 
This results in a high quality approximation of the value function, which in turn, boosts the quality of the learned value function through the fine tuning loss in \eqref{eqn:finetune_loss}.  

\subsection{Safety Assurances for the Learned Value Function} \label{sec:verification}
In general, neural network can make errors and, as such, the safe set provided by the learned value function is only a candidate safe set -- there might be states in the learned safe set that steer the system into the failure region.
To overcome this challenge, we leverage the formal verification method proposed in 
\cite{lin2024verification} that uses conformal prediction to determine a probabilistic safe set given a candidate value function $V_\theta$ and the induced safety policy, $u_{\theta}$. 

The overall idea is to provide a high-confidence bound $\delta$ on the value function error. 
The corrected value function is then used to compute the BRT and the safe set.
Specifically, the method requires users to specify a confidence parameter $\beta \in (0,1)$ and a safety violation parameter $\epsilon \in (0,1)$. It then computes the error bound $\delta$ using conformal prediction to ensure that with at least $1-\beta$ confidence, the probability of safety violation within the super-$\delta$ level set of $V_\theta$ is upper-bounded by $\epsilon$:
\begin{equation}
    \underset{x \in {\mathcal{S}_{\delta}}}{\mathbb{P}}( V(x, 0) \leq 0) \leq \epsilon,
\end{equation}
where $\mathcal{S}_{\delta}=\left\{x: x \in \mathcal{X}, V_{\theta}(x,0)>\delta \right\}$ and $V(x,0)$ is the underlying ground-truth value function. In other words, the probability of a state within the super-$\delta$ level set of $V_{\theta}$ being actually unsafe is at most $\epsilon$. 
Thus, $\mathcal{S}_{\delta}$ represents a high-confidence estimate of the safe set.

\begin{figure*}[t!]
		\centering
         {\includegraphics[width=1.0\textwidth]{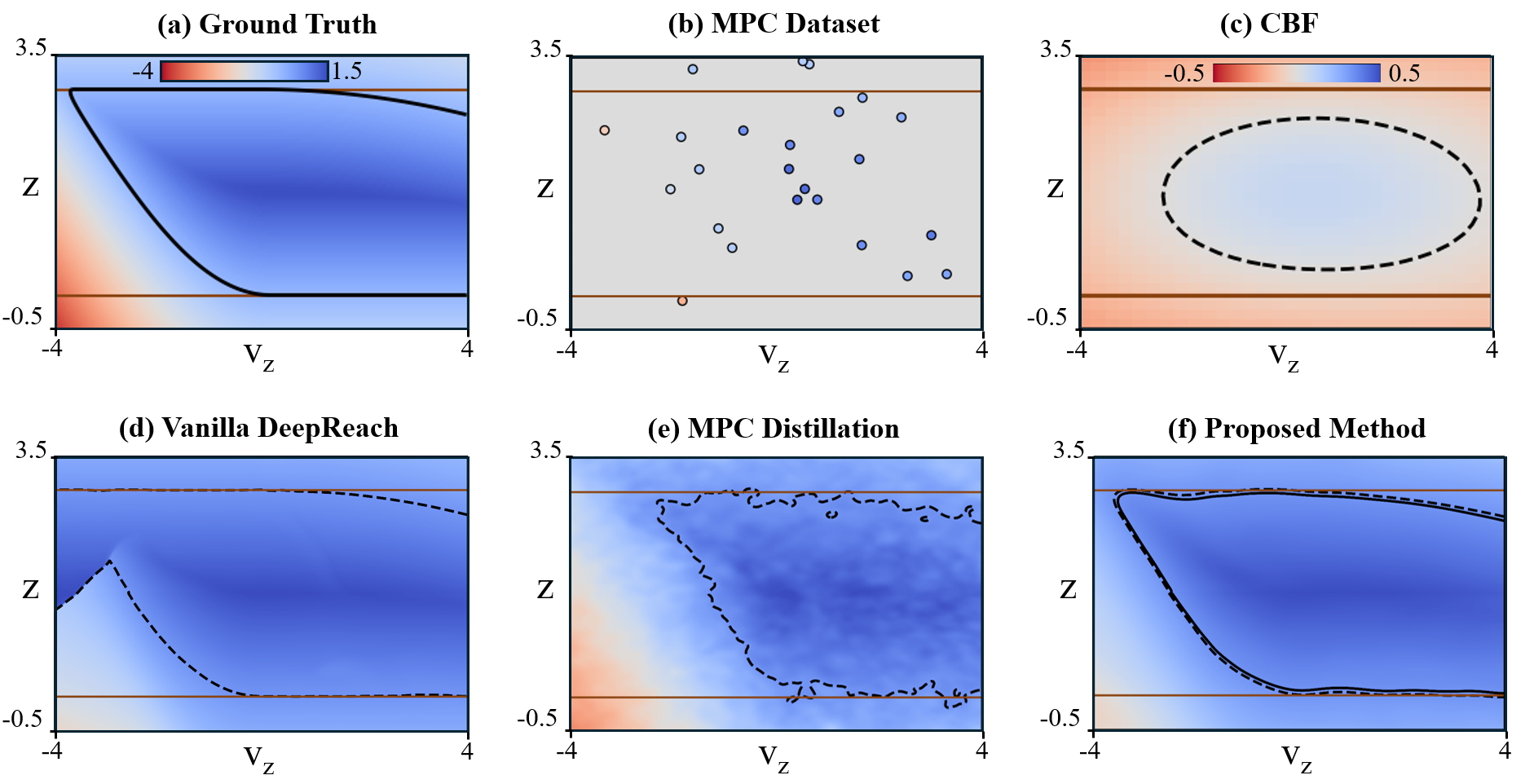}}
 	\caption{\small{Parameterized Vertical Drone: Value function slices at $K=12$. The brown lines represent the ground and the ceiling -- the failure set in this case. In (a), the solid black lines are the contours of ground-truth safe sets. In (b), all MPC data samples with $K\in (11,12)$ used for the proposed approach are shown. (c) illustrates the learned safe set using the neural CBF approach. 
    % It uses a separate color bar since the value function in CBF is differently defined than in HJ reachability. 
    For (d), (e), and (f), the dashed contours illustrate the learned safe sets and the solid contours represent the verified safe sets. Note that the black solid contours are missing in (d) and (e) since Vanilla DeepReach and MPC Distillation have an empty verified safe set.}}
    \vspace{-1.5em}
 	\label{fig:PVD_BRTs}
\end{figure*}

%% file: results.tex
We now evaluate the benefits of including MPC-based guidance when learning reachability solutions.
We consider four different case studies: a 2D vertical drone system, a 13D nonlinear quadrotor system, a 7D high-speed F1Tenth car system, and a 40D publisher-subscriber system.
Each of these systems presents unique challenges for reachability analysis due to their dimensionality, complex dynamics, and/or complex geometry of the failure set.

\subsection{Baselines}
We compare the safe set obtained via the proposed method with the following baselines:
\begin{itemize}
    \item \textbf{Vanilla DeepReach} \cite{singh2024imposing}: A NN-based safety value function obtained using DeepReach with exact boundary condition imposition (self-supervision only).
    \item \textbf{MPC Distillation}: A NN-based safety value function distilled from a dense MPC dataset (supervision only).
    \item \textbf{Neural CBF}: A NN-based control barrier function (CBF) learned using the approach proposed in \cite{dawson2022learning,pmlr-v164-dawson22a}.
    \item \textbf{Ground Truth Value Function}: Wherever possible, we compute the ground-truth value function by solving the HJB-VI numerically using the OptimizedDP toolbox \cite{bui2022optimizeddp}. Note that this is computation is feasible only up to 5D-6D systems.    
\end{itemize}
 Each of the value functions corresponds to an implicit safe policy, as defined in Eqn~(\ref{eq:safe_control}). We additionally compare the efficacy of these learned safe policies in maintaining system safety.

\subsection{Evaluation Metrics}
Our key evaluation metric is the volume of the (verified) safe set, denoted as $\mu_{\epsilon}$, which we refer to as the \textbf{recovered volume}. 
For all value functions, the verified safe set is obtained by the verification procedure outlined in Sec. \ref{sec:verification}, with $\epsilon=10^{-3}$ and $\beta = 10^{-16}$.
This corresponds to obtaining a safe set, $\mathcal{S}_{\delta}$ with a $99.9\%$ safety rate.
To estimate the volume of the safe set over the state space of interest, we compute the likelihood of a uniformly sampled state lying in the set:
\begin{equation}
    \mu_{\epsilon} \approx \frac{\sum_{i=1}^M \mathbbm{1}(x_i \in \mathcal{S}_{\delta})}{M}, x_i \sim Uniform(\mathcal{X}).
\end{equation}
We choose a large value of $M = 1 \times 10^6$ to attain samples from a significant portion of the state space.

For low dimensional problems where the ground truth is available, we also compute the Mean Squared Error (MSE) and false positive rate compared to the ground-truth value function.

\subsection{Implementation Details}
For all baselines, we employ a three-layer sinusoidal NN, with 512 neurons per layer. The NNs are trained using the Adam optimizer with a learning rate of $\alpha = 2\times 10^{-5}$ on an NVIDIA GeForce RTX 4090 GPU. 
The sampling-based MPC method uses a time step of $\Delta=0.02s$ and 10 iterative sampling steps ($R=10$), where N=100 perturbed control sequences are generated per step. The fine-tuning loss weight $\lambda_{FT}$ is set to $100$ and the dataset refinement horizon $H_R$ is set to $0.2s$ for all experiments.

In each case study, we use the same number of training iterations across all baselines to ensure a fair comparison. However, Vanilla DeepReach and MPC Distillation do not involve a fine-tuning phase, as we observe that fine-tuning encourages overfitting in these baselines and degrades their performance. 
Detailed training parameters are provided in Table~\ref{tab:params}.

\begin{table}[t]
\centering
\input{table/param}
\caption{Detailed parameters for training.}
\label{tab:params}
\end{table}

\begin{table}[t]
\centering
\input{table/baselines}

\caption{Recovered safe set volumes and training time for different case studies. The proposed approach consistently achieves higher safe set volumes compared to the baselines. We note that the safe sets obtained by the NeuralCBF method cannot be verified directly; thus, the reported (learned) volumes are likely optimistic and would reduce further upon verification. For the vertical drone system, the volumes corresponding to $K=12$ are shown in brackets. \vspace{-.3em}}
\label{tab:times}
\end{table}
The total training time for all baselines is reported in Table~\ref{tab:times}. The additional computation cost of the proposed method mainly arises from the weight-balancing step (Line 10 in Alg.~\ref{alg:training}) and computing the data-driven loss (gradients).
\subsection{Parameterized 2D Vertical Drone}
We consider a drone navigating longitudinally along the z-axis, with its dynamics defined as:
\begin{equation}
    \dot{z} = v_z, \
    \dot{v}_z = K u - g, \
    \dot{K} = 0,
\end{equation}
where $z \in [-0.5,3.5]$ is the height, $v_z \in [-4.0, 4.0]$ denotes the vertical velocity, $g = 9.8$ is the gravitational acceleration, and $K \in [0.0, 12.0]$ is a constant control gain parameter. 
The control input $u \in [-1,1]$ corresponds to the vertical acceleration effort. The system aims to avoid crashing into the ground or the ceiling, with its failure set $\mathcal{L} = \left\{ x : |z-1.5| \geq 1.5 \right\}$. 
Our goal is to compute the safe set of the system for different values of $K$.

\begin{figure*}[t!]
    \centering
    \includegraphics[width=\linewidth]{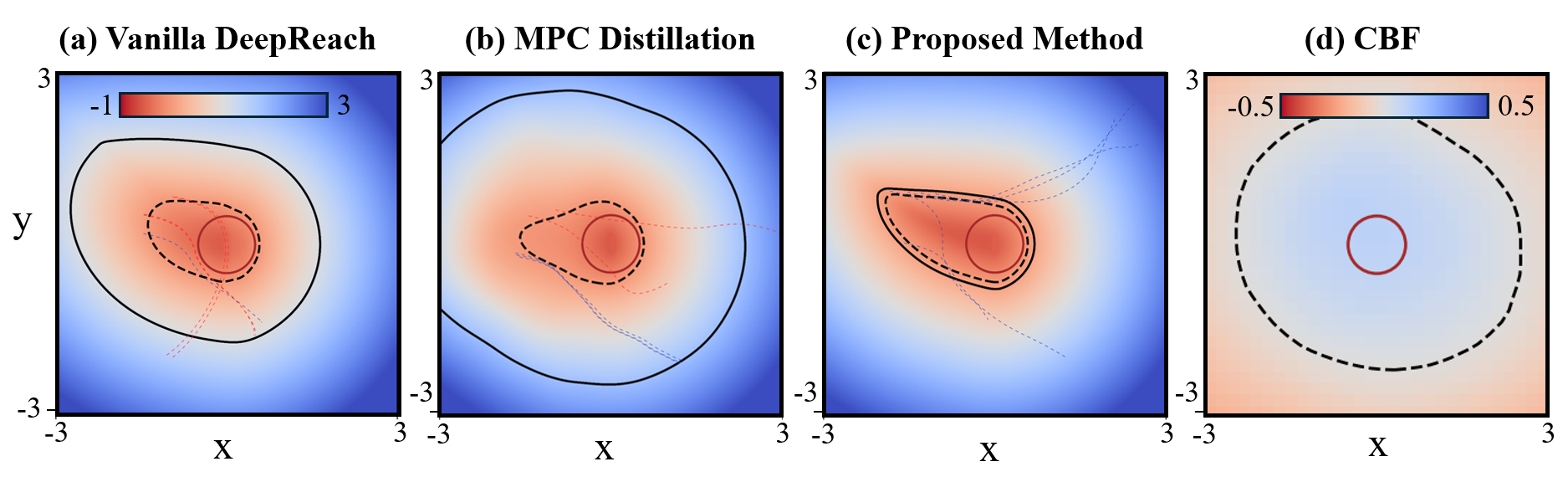}
    \caption{\small{13D Quadrotor: slices of learned value functions on the X-Y plane at $(p_z,q_\omega,q_x,q_y,q_z, v_x, v_y, v_z, \omega_x, \omega_y, \omega_z)=(0.54,  0.44, -0.45,  0.27, -0.73, 5.00, -1.07, -3.34, 3.19, -2.80,  3.43)$. The brown circles represent the cylinder obstacle (not the failure set contour due to the quadrotor’s nonzero collision radius). 
    Again, the solid contours are the verified safe sets while the dashed contours are the learned BRTs. The dashed lines are selected rollout trajectories, where red color indicates collisions.}}
    \label{fig:quadrotor_BRTs}
\end{figure*}

Since the dynamics are low-dimensional (2D state and 1D parameter), we compute the ground-truth value function through OptimizedDP \cite{bui2022optimizeddp} and use it to evaluate the mean squared error (MSE). The ground truth safe set volume is 26.22\%. To train the proposed method, we generate 300 MPC data points ($|D_{MPC}|=300$) in approximately 1s for the parameterized 2D vertical drone system.
These data points were bootstrapped instantly using the procedure discussed in Remark 2, leading to an overall dataset of size around 10K.

We first compare the performance of all baselines. 
As shown in Fig.~\ref{fig:PVD_BRTs}, Vanilla DeepReach converges to a completely non-physical solution, with an MSE of 7.81 and a dramatically overoptimistic safe set.
This particular reachability problem represents a corner case where Vanilla DeepReach suffers from instability issue, despite the simplicity of both the system dynamics and boundary conditions, and results in a verified safe set volume of zero.  
On the other hand, the distilled value function, which minimizes data-driven loss on a dense dataset with roughly 10M MPC samples, attains a lower MSE of $0.41$. 
Although the distilled value function is closer to the ground truth, it exhibits a noisy value profile due to the absence of PDE loss. 
Consequently, the learned value function and the induced safety policy suffer from inaccuracies, resulting in a recovered safe volume of zero, demonstrating that the MPC method alone is not sufficient to obtain a high-quality value function despite a dense dataset.
In contrast, the proposed method shows a significant improvement with a small MPC dataset ($|D_{MPC}|=300$), achieving a low MSE of $0.009$ and a recovered volume of $24.62\%$.

Since the Neural CBF baseline learns a converged safe set, we only compare this baseline with the proposed approach for $K=12$. For $K=12$, the safe set converges within $T=1.2~s$, enabling a fair comparison. 
At $K=12$, the learned safe set volume from Neural CBF is $31.28\%$, whereas our method achieves a significantly larger recovered volume of $56.51\%$.
This is also evident from the respective safe set boundaries (Fig.~\ref{fig:PVD_BRTs}(c) and (f)).

\begin{figure}[H]
    \centering
    \includegraphics[width=0.99\linewidth]{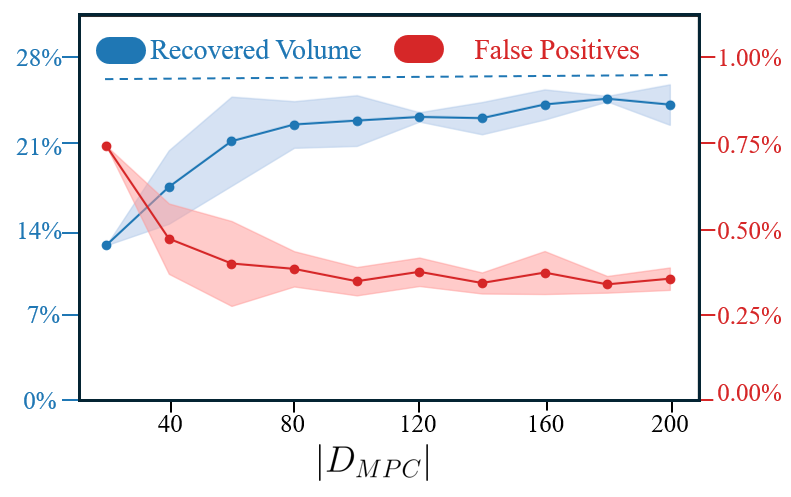}
    \caption{\small{The x-axis shows the size of non-bootstrapped dataset. The blue line represents the mean recovered volumes along with their standard deviation across five random seeds, whereas the red line shows the mean false positive rates.}}
    \label{fig:dataset_size}
\end{figure}

\textit{Ablation study: effect of MPC dataset size.} To assess the effect of dataset size, we train the value function with several sparse datasets. The number of initial states used to generate the MPC dataset ranges from 20 to 200, resulting in approximately 700 to 7000 data samples throughout the time horizon after bootstrapping. Each dataset takes less than a second to generate. 
For each dataset size, we train a 3-layer sinusoidal network with 128 neurons per layer using 5 different random seeds.  
As illustrated in Fig.~\ref{fig:dataset_size}, the recovered volume quickly converges to around $24\%$ with $N_{MPC} \approx 160$. This result confirms that the proposed method exhibits a high data efficiency and learns accurate value function with minimal MPC guidance. This characteristic becomes increasingly important as system dimensionality increases and the data available becomes more sparse.
The same trend is observed in the false positive rate (evaluated using ground truth value function).

\textit{Ablation study: effect of MPC dataset refinement.} Given the MPC dataset is inherently suboptimal and noisy, we investigate the impact of the iterative dataset refinement step on the algorithm's performance. To this end, we evaluate the proposed method without dataset refinement, which results in a higher MSE of $0.2285$ and a lower recovered volume of $20.04\%$.
All metrics deteriorate compared to the case where the MPC dataset is refined periodically, highlighting the utility of the refinement step to reduce the suboptimality of the MPC guidance. 
\begin{figure*}[t!]
\centering
\subfloat{\includegraphics[width=0.248\textwidth]{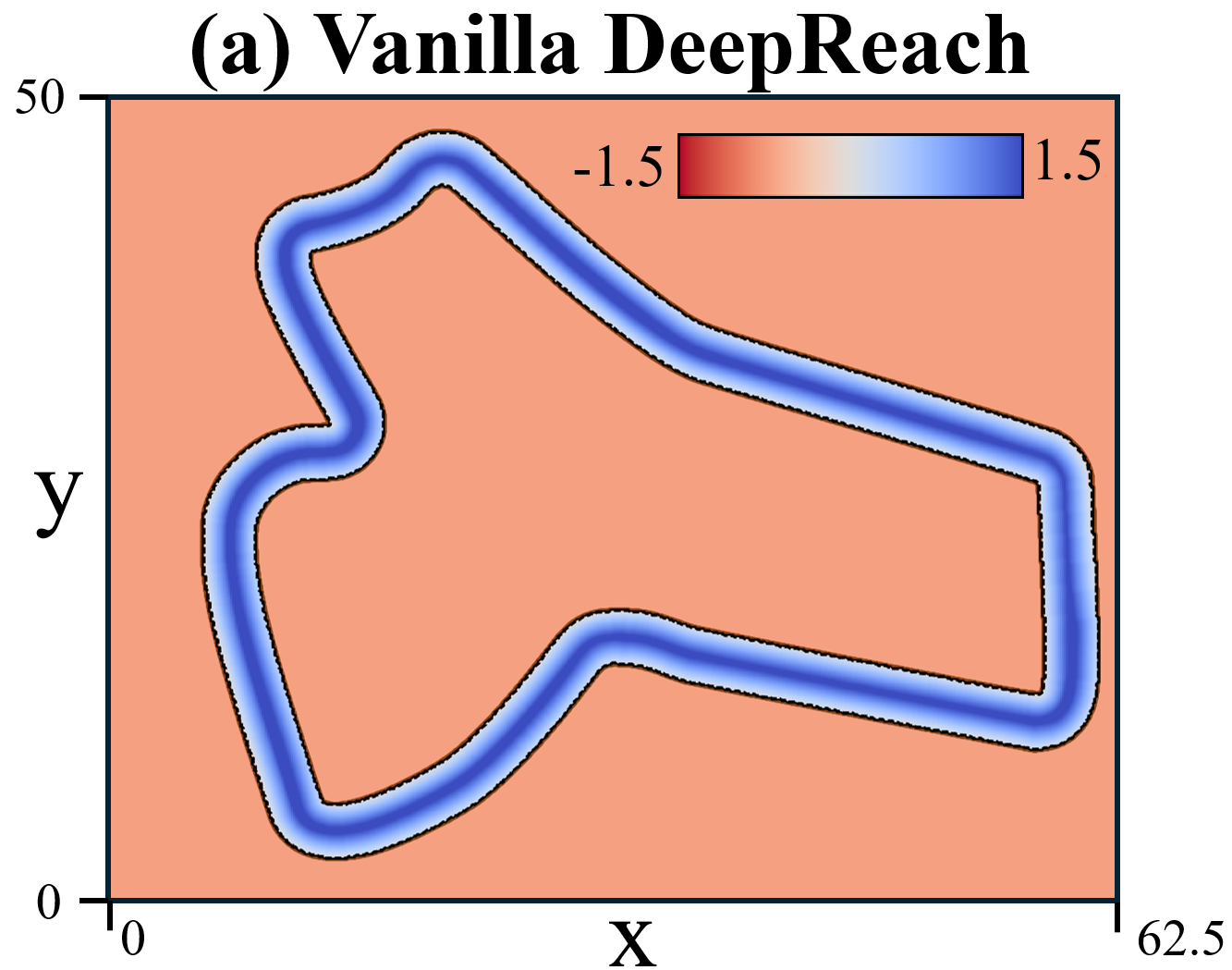} \label{fig:f10_BRTs1}}
\subfloat{\includegraphics[width=0.248\textwidth]{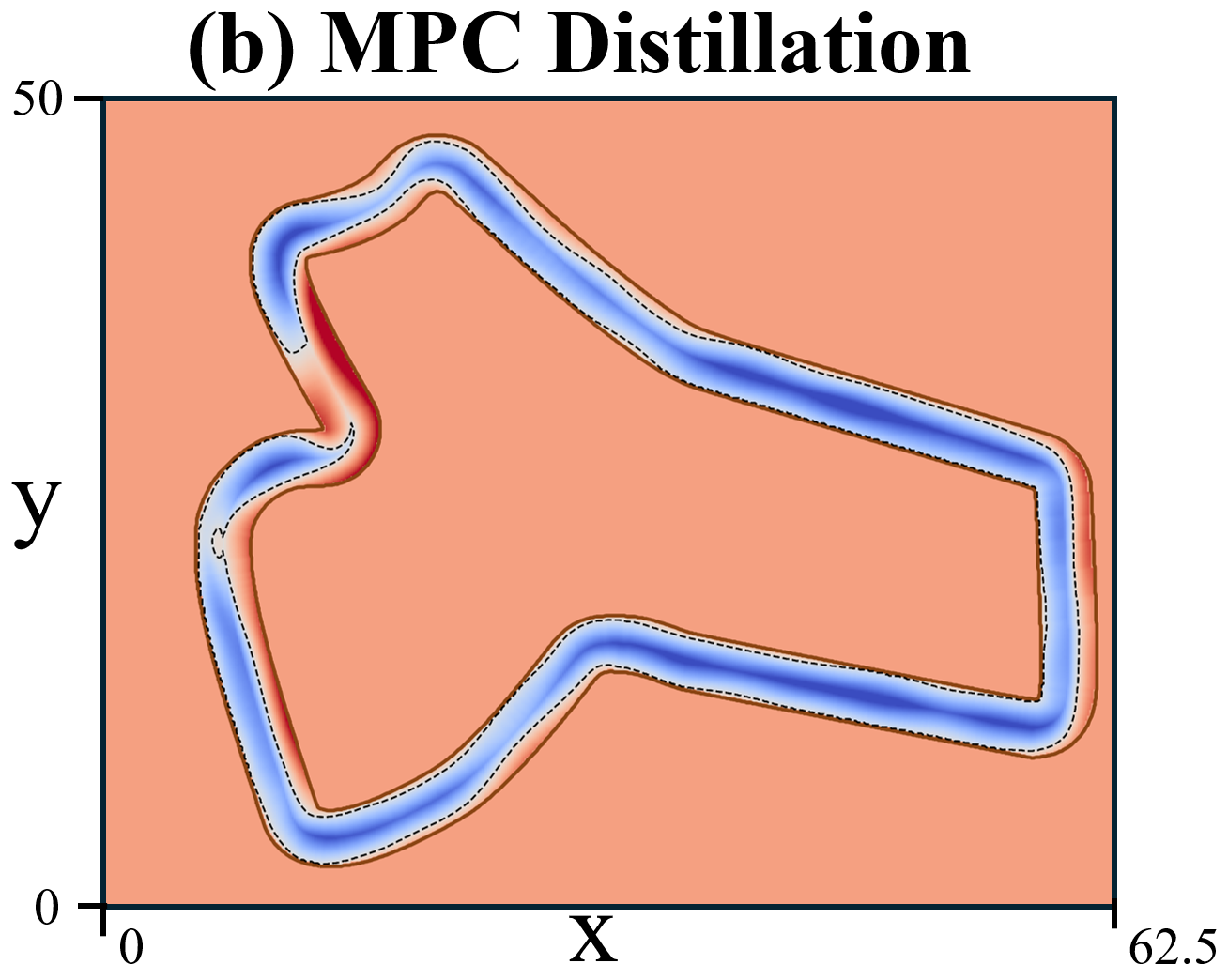} \label{fig:f10_BRTs2}}
\subfloat{\includegraphics[width=0.248\textwidth]{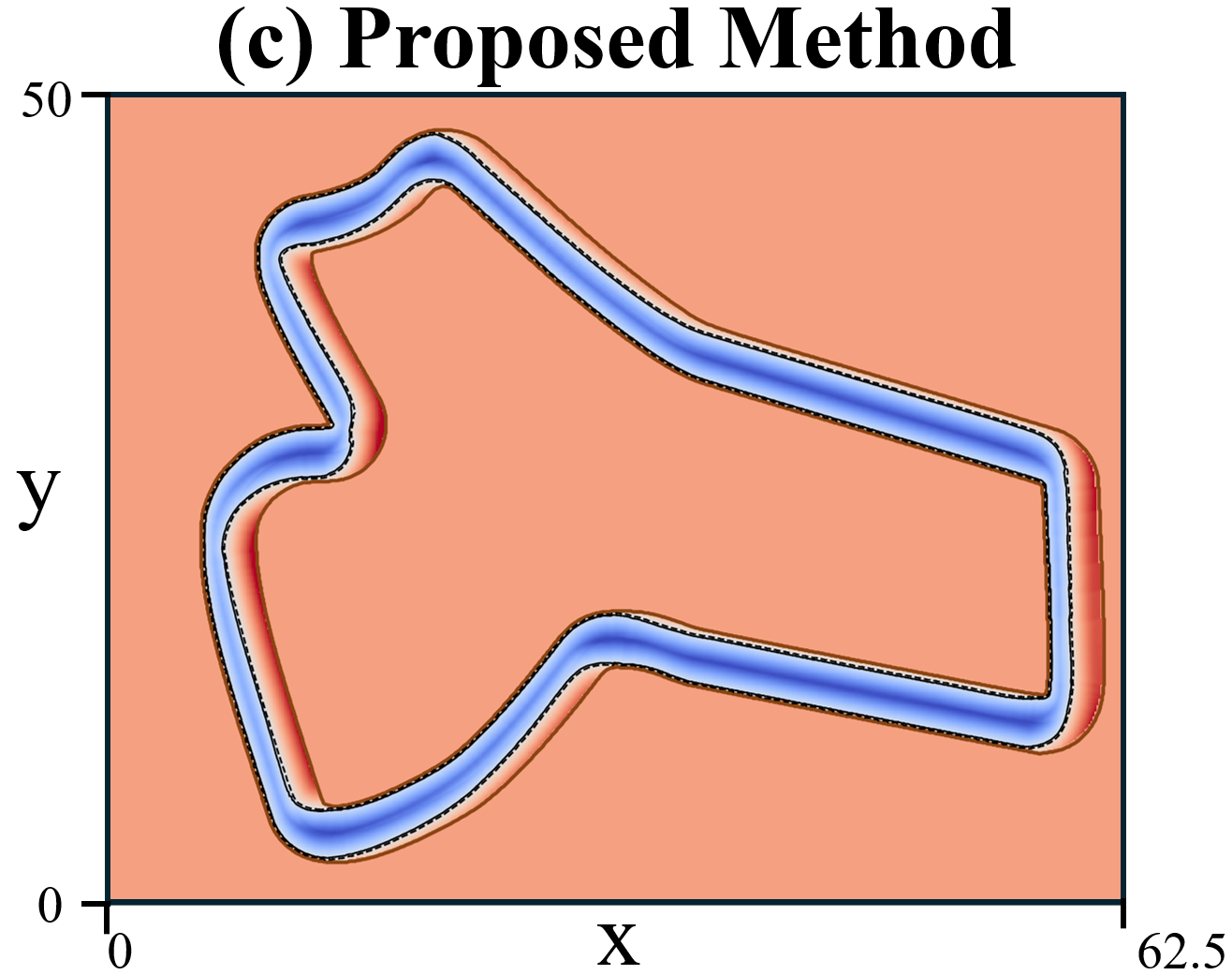} \label{fig:f10_BRTs3}}
\subfloat{\includegraphics[width=0.248\textwidth]{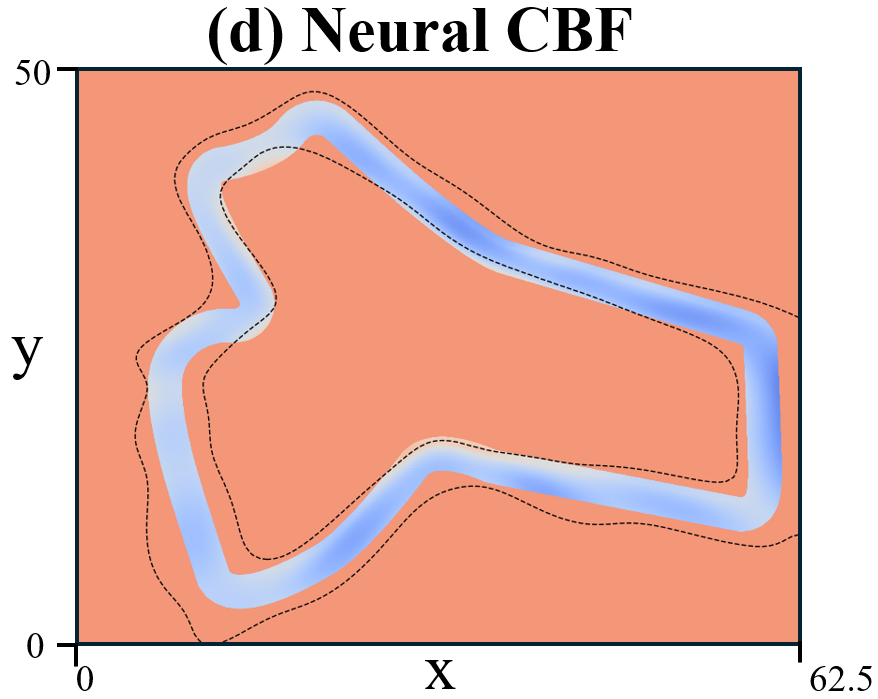} \label{fig:f10_BRTs4}}
\caption{\small{F1Tenth: X-Y Plane Slices of the Learned Value Functions. The remaining state variables are $(\phi , v , \theta_{yaw} , \omega_{yaw} , \beta_{slip})=(0,8,0,0,0)$. The track boundaries are shown as the brown contours. The black solid contours are the verified safe sets while the dashed contours are the learned BRTs. Note that the black solid contours are missing in (a), (b), and (d) since these baselines have an empty verified safe set. On the other hand, the black solid contour and dashed contour overlap significantly for the proposed method, with $\delta=0.12$, highlighting the high accuracy of the learned safety value function.}}
\end{figure*}

\subsection{13D Quadrotor System}
We next evaluate the proposed method and the baselines on a high-dimensional reachability problem involving an extremely agile quadrotor system. 
The state of the drone is represented as $x=[p_x,p_y,p_z,q_\omega,q_x,q_y,q_z, v_x, v_y, v_z, \omega_x, \omega_y, \omega_z]$, where $(p_x, p_y, p_z) \in [-3,3]^3$ denote the position and $(v_x,v_y,v_z) \in [-5,5]^3$ denote the linear velocities. $(q_\omega,q_x,q_y,q_z) \in [-1,1]^4$ represent the quaternion and $(\omega_x, \omega_y, \omega_z)\in [-5,5]^3$ are angular velocities. 
The control inputs consist of a collective thrust and angular acceleration efforts $u=[F, \alpha_x, \alpha_y, \alpha_z]$, where $F \in [-20,20]$, $\alpha_x, \alpha_y \in [-8,8]$, and $\alpha_z \in [-4,4]$. The quadrotor is modeled as a disk with a radius of $0.17$~m. 
The safety function $l(x)$ is defined as the signed distance from the disk to an infinitely long cylinder with a radius of 0.5~m centered at the origin. 
Consequently, the failure set $\mathcal{L}$ comprises all states where the quadrotor collides with the obstacle. 
As a reference, the failure set volume is 3.09\%. A detailed description of $l(x)$ and the quadrotor dynamics is provided in the Appendix~\ref{appendix:quadrotor_dynamics}. 

Once again, Vanilla DeepReach learns an overly optimistic value function. 
The inaccuracies in the value function lead to a low recovered safe volume of $71.26\%$. 
In contrast, the proposed method, trained with 1M data samples, significantly outperforms Vanilla DeepReach by achieving a higher recovered volume of $93.69\%$, highlighting the utility of MPC-based guidance. 
The MPC Distillation baseline, despite being trained on 15M data samples, leads to the lowest recovered volume of $42.99\%$.
This is because the dataset remains too sparse to provide adequate interpolation in this high-dimensional problem, highlighting the utility of incorporating PDE loss during the value function learning.
The Neural CBF baseline attains a learned volume of $51.7\%$ and does not have a recovered volume since the verification method is not applicable. 
Overall, the proposed method yields the best safe controller along with the largest verified safe set. 
Note that although all baselines appear to achieve a substantial verified safe set, the performance gaps remain significant given the small failure set and the agility of the quadrotor, as better illustrated in Fig.~\ref{fig:quadrotor_BRTs}.

\textit{Ablation studies.}
We conducted ablation studies on the MPC dataset size and found that the proposed method can synthesize high-fidelity BRTs using relatively sparse data (Table~\ref{Table:dataset_size}). Additionally, we observed that removing the time curriculum reduced the recovered volume to zero, whereas omitting either of the other two stages had a minimal impact (Table~\ref{tab:ablation_training_stages}). These results highlight the critical role of the time curriculum in solving complex reachability problems despite its additional computational cost, while the other two stages are optional and primarily enhance stability further. 

\begin{table}[h]
\centering
\resizebox{\columnwidth}{!}{%
\begin{tabular}{|cc|cc|cc|cc|}
\hline
\multicolumn{2}{|c|}{\begin{tabular}[c]{@{}c@{}}\textbf{$|D_{MPC}|$}\\\textbf{Quadrotor/F1}\end{tabular}} & 
\multicolumn{2}{c|}{\begin{tabular}[c]{@{}c@{}}\textbf{Bootstrapped}\\\textbf{Dataset Size}\end{tabular}} & 
\multicolumn{2}{c|}{\begin{tabular}[c]{@{}c@{}}\textbf{Dataset}\\\textbf{Generation Time}\end{tabular}} & 
\multicolumn{2}{c|}{\begin{tabular}[c]{@{}c@{}}\textbf{Recovered}\\\textbf{Volume}\end{tabular}} \\
\hline
5k & 150k & 35k & 1.8M & 20~s & 6~m & 89.8\% & 0.0\%\\
15k & 210k & 110k & 2.6M & 1.5~m & 8~m & 91.7\% & 53.4\% \\
30k & 300k & 230k & 3.3M & 2.5~m & 10~m & 93.7\% & 76.3\% \\
50k & 360k & 390k & 4.5M & 4~m & 13~m & 93.8\% & 76.0\% \\
100k & 450k & 750k & 5.9M & 6~m & 15~m & 93.5\% & 74.2\%\\
\hline
\end{tabular}%
}
\caption{Dataset statistics for Quadrotor (Left) and F1Tenth (right).}
\label{Table:dataset_size}
\end{table}

\begin{table}[h]
\centering
\begin{tabular}{|l|c|c|c|}
\hline
\textbf{Training Configuration} & \textbf{Vertical Drone} & \textbf{Quadrotor} & \textbf{F1Tenth} \\
\hline
Full Training (All Stages)          & 24.62\% & 93.69\% & 76.08\% \\
\hline
Without Fine-tuning                 & 24.37\% & 93.18\% & 66.53\% \\
\hline
Without Pretraining                 & 24.00\% & 93.87\% & 65.61\% \\
\hline
Without Time Curriculum         & 23.50\% & 0.00\%  & 0.00\%  \\
\hline
\end{tabular}
\caption{Effect of training stages on recovered volume.}
\label{tab:ablation_training_stages}
\end{table}

\subsection{7D F1Tenth}
To further assess the robustness and accuracy of our method, we introduce an extremely challenging reachability problem with both stiff dynamics and a complex boundary condition: the F1Tenth racing car problem. 
The system is tasked with navigating a high-speed RC car on a track while avoiding the curbs. The failure set $\mathcal{L}$ consists of all states where the car is outside the track, and $l(x)$ is defined as the signed distance function to the edge of the track. 

The state variables of F1Tenth is given by $x=(p_x, p_y, \phi , v , \theta_{yaw} , \omega_{yaw} , \beta_{slip}) \in [0,62.5] \times [0,50]\times [-0.4189, 0.4189] \times [0, 8] \times [-\pi, \pi] \times [-5, 5] \times [-0.8, 0.8]$, where $( p_x, p_y )$ are the global coordinates of the vehicle, $\phi$ is the front-wheel steering angle, $v$ is the velocity, $\theta_{yaw}$ is the yaw angle, $\omega_{yaw}$ is the yaw rate, and $\beta_{slip}$ denotes the slip angle at the vehicle's center. The control inputs are the rate of the front-wheel steering angle and the longitudinal acceleration, denoted as $u=[\dot{\phi}, a] \in [-3.2,3.2] \times [-9.51, 9.51]$. The longitudinal acceleration is further capped when the current velocity $v$ is high, and the system switches to kinematic mode when the velocity is too low. A detailed description of the system dynamics is provided in the Appendix~\ref{appendix:f1tenth}.

Given the hybrid dynamics and high agility of the vehicle, learning an accurate safety value function is challenging. 
This challenge is further exasperated by the complex geometry of the failure set and the large state space of interest.
Both Vanilla DeepReach and the MPC Distillation approach fail completely to recover any verified safe set volume in this case. Vanilla DeepReach suffers again from the instability issue, causing the learned value function to not evolve through the time curriculum and instead converges to $l(x)$, as shown in Fig.~\ref{fig:f10_BRTs1}.
In this case study, the Neural CBF baseline performs the worst, producing a completely nonphysical safe set that intersects heavily with the failure set.
In contrast, our method significantly outperforms all baselines by achieving an impressive recovered volume of $76.08\%$ with 3M data samples.

We further roll out the obtained safety policy, as well as the sampling-based MPC policy, from different initial states and quantify the empirical safe volume under these policies. 
We compare the two policies for $v>6$ as the race car primarily operates at a high speed. The policy learned by our method demonstrates a more aggressive (fallback) driving strategy, achieving a safe volume of $76.87\%$, compared to $69.38\%$ for the MPC policy, highlighting the utility of HJB-VI residuals that encourage global optimality in the learned value function and the safety policy. 
In terms of online inference speed, our method computes the optimal safe control in approximately 2~ms, whereas the sampling-based MPC approach requires 39~ms. Both methods could be further accelerated by migrating to C++.

\textit{Ablation studies.}
The results of the MPC dataset size and training stage ablation studies are presented in Table~\ref{Table:dataset_size} and Table~\ref{tab:ablation_training_stages}, respectively, and yield consistent conclusions. The curriculum training remains the key training phase for the proposed approach. Notably, both computation time and dataset size scale primarily with problem complexity rather than system dimensionality, as the 7D F1Tenth Race Car problem requires more computation time and dataset samples compared to the higher-dimensional 13D quadrotor problem. This conclusion is further corroborated with a 40D publisher-subscriber system in the next case study.

\begin{figure}[h]
\centering
\subfloat{\includegraphics[width=0.25\textwidth]{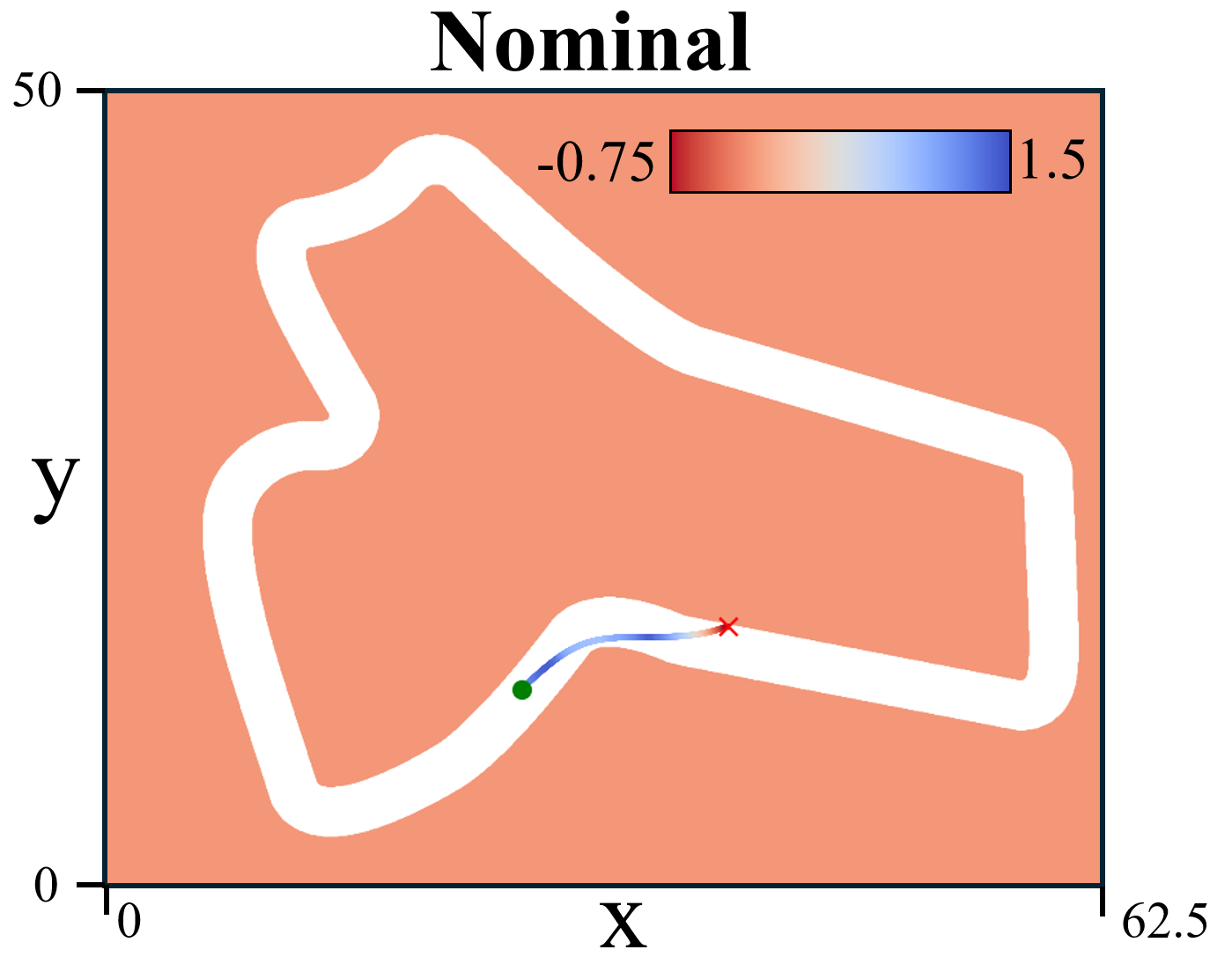}\label{fig:nominal_1}}
\subfloat{\includegraphics[width=0.25\textwidth]{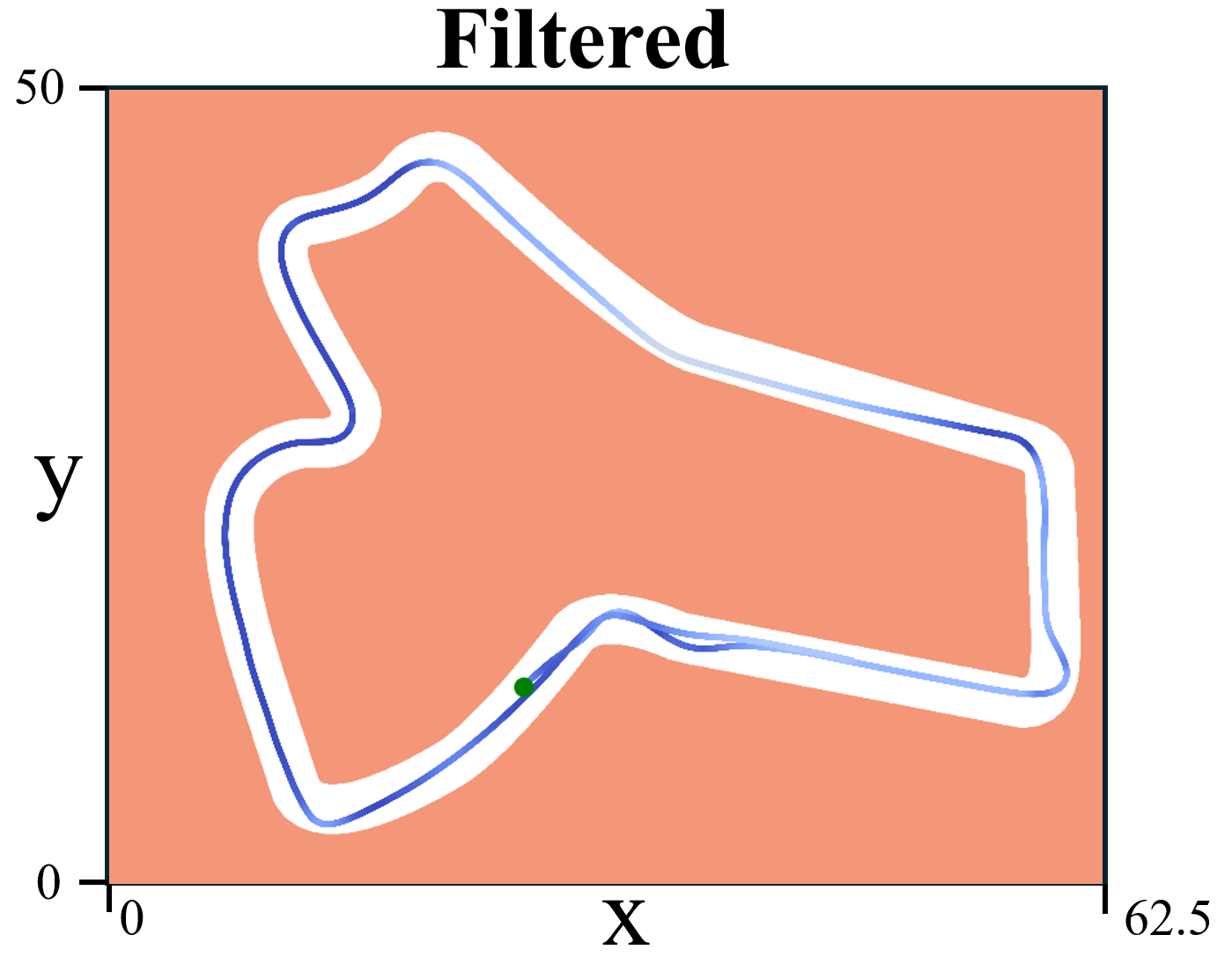} \label{fig:filter_1}} \\
\caption{\small{Trajectories under the nominal controller and the safety filter with $\gamma=1$. The green dot marks the starting points of the trajectories, while the red ``X'' denotes collisions. The filtered policy successfully completes an entire lap without collision.}}
\label{fig:safety_filter}
\end{figure}

\subsubsection*{\textbf{Safety Filtering}} 
Reachability-based safety value function can be used for safety filtering where a potentially unsafe nominal controller is minimally modified to render it certifiably safe.
The nominal controller here is an imitation policy trained using an MPC-based expert. 
We implement a smooth blending safety filter, inspired by CBF-QP, following the approach in \cite{borquez2024safety} with $\gamma=1.0$.  

To assess both performance and safety, we evaluate the nominal and filtered policies based on two metrics: collision rate and distance travelled before collision.
We evaluate the system for a horizon of $25 \, \text{s}$ and each policy is evaluated over 30 different trajectories. 
At the start of each trajectory, the F1Tenth car is initialized inside the verified safe set (as per our method). 
Fig.~\ref{fig:safety_filter} illustrates trajectories under the nominal controller and the safety filter along with the value function predictions. The nominal controller fails to keep the car on the track, whereas the filtering technique---leveraging the learned safety value function---enables the car to complete an entire lap without collision.
Quantitatively, the nominal policy achieves an average distance travelled of $36.17 \, \text{m}$ with a 100\% collision rate. The filtered policy successfully operates without any collision, achieving average distances of $187.01 \, \text{m}$. These results highlight a substantial improvement in both performance and safety when leveraging the learned value function for downstream safety filtering and control.

\subsection{Case Study: 40 Dimensional Publisher-Subscriber System}\label{case study:40D system}
To demonstrate the scalability of the proposed approach to high-dimensional systems, we apply it to a 40-dimensional ``publisher-subscriber'' system, defined as:
\begin{equation}
\begin{aligned}
    \left[\begin{array}{l}
\dot{\mathrm{x}}_0 \\
\dot{\mathrm{x}}_i
\end{array}\right] \triangleq\left[\begin{array}{cc}
a & 0 \\
-1 & a
\end{array}\right]\left[\begin{array}{l}
\mathrm{x}_0 \\
\mathrm{x}_i
\end{array}\right]+\left[\begin{array}{l}
0 \\
b
\end{array}\right] u_i+&\left[\begin{array}{c}
\alpha \sin \left(\mathrm{x}_0\right) \mathrm{x}_0^2 \\
-\beta \mathrm{x}_0 \mathrm{x}_i^2
\end{array}\right], \\
\ \ &i=1,...,39,
\end{aligned}
\end{equation}
where $x_0$ represents the publisher state, which unidirectionally influences each subscriber state $x_i$. For this study, we set $\alpha=20$, $\beta=0$, and  constrain the control input by $\left|u_i\right| \leq 1$. The target set which the agents aim to reach is given by $\mathcal{L} =\left\{x:  \frac{1}{2} (x^2_0 + x^2_i - 0.5) \leq 0 , \forall i=1,...,39\right\}$. The ground truth value function is obtainable via system decomposition; we refer interested readers to \cite{sharpless2024linearsupervisionnonlinearhighdimensional} for further system details.

We adapt the same network architecture as the previous case studies and the training parameters are provided in Table~\ref{tab:params}. Training required approximately 3 hours with a $|D_{MPC}|$ of 80K---roughly the same computational burden as the 13D quadrotor system despite the tripled dimensionality. This further supports our claim that the computational burden of the proposed method scales primarily with problem complexity rather than system dimensionality.

Vanilla DeepReach produced a value function with an MSE of 11.27 and recovered only 38.21\% of the ground truth BRT volume. In contrast, our method achieved a significantly more accurate value function with an MSE of 0.0062 and recovered 97.14\% of the ground truth BRT volume (see Fig.~\ref{fig:40D system}), demonstrating its effectiveness in high-dimensional settings.
\begin{figure}[h!]
  \centering
  \vspace{-1em}
\includegraphics[width=0.95\linewidth]{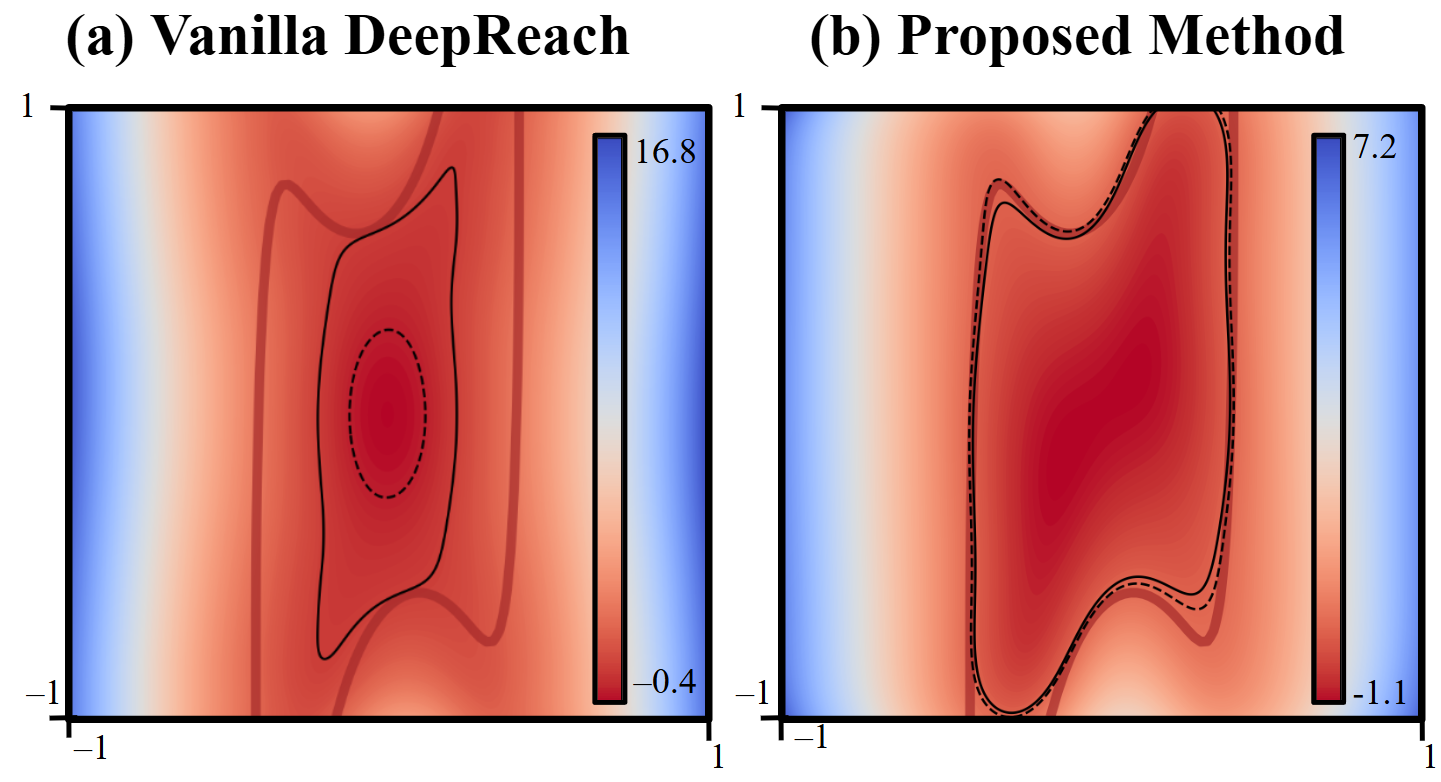}
\vspace{-0.5em}
  \caption{Value function slices for the 40D system. The brown contours represent the ground-truth BRT slices. The dashed lines indicate the learned BRTs while the solid lines are the verified safe set contours.\label{fig:40D system}}
  \label{fig:40D system}
\end{figure}

%% file: table/param.tex
\begin{tabular}{|c|c|c|c|c|}
\toprule
\textbf{Parameter} & \textbf{Vertical} & \textbf{Quad-} & \textbf{F1Tenth} & \textbf{40D Publisher-} \\
                  & \textbf{Drone}    & \textbf{rotor} &                  & \textbf{Subscriber}     \\
\hline 
\textbf{$T$} & 1.2 & 1.0 & 1.0 & 1.0 \\
\textbf{$\mathcal{N}_{PDE}$} & 65K & 65K & 65K & 65K \\
\textbf{$\mathcal{N}_{MPC}$} & 10K & 10K & 10K & 10K \\
Pretraining & 20K & 20K & 20K & 20K \\
Curriculum & 40K & 100K & 200K & 120K \\
Fine-Tuning & 10K & 10K & 10K & 10K \\
\bottomrule
\end{tabular}

%% file: table/baselines.tex
\begin{tabular}{|c|c|c|c|}
\toprule
\textbf{} & \textbf{Baseline} & \textbf{Recovered (\%)} & \textbf{Time [h]} \\ 
\hline
\multirow{3}{*}{\textbf{Vertical Drone}} 
    & Vanilla & 0.00 & 0.05 \\ 
    & Distillation & 0.00 & 0.1 \\ 
    & Neural CBF & NA (31.28) & 0.3 \\
    & Proposed & 24.62 (56.51) & 0.1 \\ 
\hline
\multirow{3}{*}{\textbf{Quadrotor}} 
    & Vanilla & 71.26 & 1.6 \\ 
    & Distillation & 42.99 & 2.5 \\ 
    & Neural CBF & 51.70 & 6.0 \\
    & Proposed & 93.69 & 3.0 \\ 
\hline
\multirow{3}{*}{\textbf{F1Tenth}} 
    & Vanilla & 0.00 & 2.0 \\ 
    & Distillation & 0.00 & 4.5 \\ 
    & Neural CBF & 167.14 & 20.0 \\ 
    & Proposed & 76.08 & 5.0 \\ 
\hline
\multirow{3}{*}{\textbf{Publisher-Subscriber}} 
    & Vanilla & 38.21 & 1.3 \\  
    & Proposed & 97.14 & 2.8 \\ 
\bottomrule
\end{tabular}

%% file: conclusion.tex
In this work, we propose a framework that efficiently generates approximate value function datasets using a sampling-based MPC approach and integrates these data labels to improve state-of-the-art learning-based reachability methods. 
By leveraging data-driven supervision alongside PDE-based residuals, our method enhances the accuracy and scalability of reachability analysis for high-dimensional systems.
Through several case studies on challenging reachability problems, we demonstrate that our method consistently outperforms existing approaches, achieving better accuracy and faster convergence.

\section{Limitations}
While the results are promising, several open questions remain. One key challenge is understanding how the distribution of data across the temporal-spatial domain affects the training performance -- specifically, what sampling strategies could maximize learning efficiency. Developing importance sampling techniques, which further improve data efficiency by prioritizing informative samples, would be an interesting future direction. 

Additionally, this work focuses on safety problems, where value functions typically converge over short horizons, and has not explored long-horizon problems (e.g., reach-avoid problems).
It would be interesting to combine the proposed approach with alternative neural PDE solution methods, such as sequence-to-sequence learning \cite{mattey2022novel}, to effectively synthesize long-horizon PDE solutions.
It would also be interesting to consider alternative optimal control methods to provide guidance for learning safety value function.

Beyond these algorithmic considerations, our approach does not currently account for disturbances and dynamics uncertainty, which are critical in ensuring safety in uncertain and adversarial settings. Extending the framework to incorporate robust safety under disturbances presents a non-trivial challenge and is an important direction for future work.

%% file: appen.tex
\subsection{Quadrotor}
\subsubsection{Dynamics}\label{appendix:quadrotor_dynamics}
The quadrotor dynamics is given as follows:
\begin{equation}
    \begin{aligned}
    \dot{p_x}& = \nu_x, \\
    \dot{p_y} &= v_y, \\
    \dot{p_z} & = v_z,  \\
    \dot{q}_{\omega} & = - (\omega_x \cdot q_x)/2 - (\omega_y \cdot q_y)/2 - (\omega_z \cdot q_z)/2, \\
    \dot{q}_{x} & = (\omega_x \cdot q_w)/2 + (\omega_z \cdot q_y)/2 - (\omega_y \cdot q_z)/2, \\
    \dot{q}_{y} & = (\omega_y \cdot q_w)/2 - (\omega_z \cdot q_x)/2 + (\omega_x \cdot q_z)/2, \\
    \dot{q}_{z} & = (\omega_z \cdot q_w)/2 + (\omega_y \cdot q_x)/2 - (\omega_x \cdot q_y)/2, \\
    \dot{v}_x & =     CT   \cdot  ( 2 \cdot q_w \cdot q_y + 2 \cdot q_x \cdot q_z )  F  /m, \\
    \dot{v}_y & =  CT   \cdot ( -2 \cdot q_w \cdot q_x + 2 \cdot q_y \cdot q_z )  F /m, \\
    \dot{v}_z & =   Gz -  CT   \cdot  ( 2 \cdot q_x^2 + 2 \cdot q_y^2 - 1  )  F /m,  \\
    \dot{\omega}_{x} & =  \alpha_x - \frac{5}{9}  \omega_y \cdot \omega_z,  \\
    \dot{\omega}_{y} & =  \alpha_y + \frac{5}{9}  \omega_x \cdot \omega_z,  \\
    \dot{\omega}_{z} & =  \alpha_z, \\
    \end{aligned}
\end{equation}
where $CT=1$ is the lifting coefficient and $m=1 \ kg$ is the mass. $(p_x, p_y, p_z)$ denotes the position and $(v_x,v_y,v_z)$ denotes the linear velocities. $(q_\omega,q_x,q_y,q_z) \in [-1,1]^4$ represents the quaternion and $(\omega_x, \omega_y, \omega_z)$ are the angular velocities. 

\subsubsection{Boundary condition}\label{appendix:quadrotor_lx}
The safety function $l(x)$ for the quadrotor case study is computed as:
\begin{equation*}
    \begin{aligned}
     \mathbf{v}^{n}&=\mathbf{q}\cdot \mathbf{e}_3 \cdot \bar{\mathbf{q}}, \\
     d_x&=\frac{r_a^2 p_x^2 \nu_z^2}{p_x^2 \nu_x^2 + p_x^2 \nu_z^2 + 2 p_x p_y \nu_x \nu_y + p_y^2 \nu_y^2 + p_y^2 \nu_z^2}, \\
     d_y&=\frac{r_a^2 p_y^2 \nu_z^2}{p_x^2 \nu_x^2 + p_x^2 \nu_z^2 + 2 p_x p_y \nu_x \nu_y + p_y^2 \nu_y^2 + p_y^2 \nu_z^2}, \\
     l(x)&= max(\sqrt{x^2+y^2} - \sqrt{ d_x+ d_y}, 0 ) -r_o,\\
    \end{aligned}
\end{equation*}
where $\mathbf{q}=q_\omega + q_x \cdot i + q_y \cdot j +q_z \cdot k $, $\bar{\mathbf{q}}=q_\omega - q_x \cdot i - q_y \cdot j -q_z \cdot k $, $\mathbf{e}_3=[0,0,1]$, $r_a=0.17~m$ is the radius of the quadrotor, and $r_o=0.5~m$ denotes the radius of the obstacle. Essentially, $l(x)$ represents the signed distance function from a disk (i.e., the quadrotor) to the cylindrical obstacle centered at the origin with an infinite length along the z-axis.
\subsection{F1Tenth}\label{appendix:f1tenth}
The F1Tenth has a hybrid dynamics with a kinematic mode and a dynamic mode. It's in kinematic mode if the magnitude of speed is less than 0.5. Otherwise, the dynamic mode will be employed. We now present the dynamics equation for each mode.

\subsubsection{Kinematic Model Dynamics (\( |v| < 0.5 \))}
For low velocities (\( |v| < 0.5 \)), the kinematic model is used.

The kinematic-mode dynamics are:
\[
\mathbf{f} = \begin{bmatrix}
v \cos(\theta_{yaw}) \\
v \sin(\theta_{yaw}) \\
\dot{\phi} \\
a \\
\frac{v}{l_r + l_f} \tan(\phi) \\
\frac{a}{l_r + l_f} \tan(\phi) + \frac{v}{(l_r + l_f) \cos^2(\phi)} \dot{\phi} \\
0
\end{bmatrix}.
\]

\subsubsection{Dynamic Model Dynamics (\( |v| \geq 0.5 \))}
For higher velocities (\( |v| \geq 0.5 \)), the dynamic model is used. 

The dynamic-mode dynamics are:
\[
\mathbf{f} = \begin{bmatrix}
v \cos(\theta_{yaw} + \beta_{slip}) \\
v \sin(\theta_{yaw} + \beta_{slip}) \\
\dot{\phi} \\
a \\
\omega_{yaw} \\
-\frac{\mu m}{v I (l_r + l_f)} \left( l_f^2 C_{Sf} (g l_r - a h) + l_r^2 C_{Sr} (g l_f + a h) \right) \omega_{yaw} \\
+ \frac{\mu m}{I (l_r + l_f)} \left( l_r C_{Sr} (g l_f + a h) - l_f C_{Sf} (g l_r - a h) \right) \beta_{slip} \\
+ \frac{\mu m}{I (l_r + l_f)} l_f C_{Sf} (g l_r - a h) \phi, \\
\left( \frac{\mu}{v^2 (l_r + l_f)} \left( C_{Sr} (g l_f + a h) l_r - C_{Sf} (g l_r - a h) l_f \right) - 1 \right) \omega_{yaw} \\
- \frac{\mu}{v (l_r + l_f)} \left( C_{Sr} (g l_f + a h) + C_{Sf} (g l_r - a h) \right) \beta_{slip} \\
+ \frac{\mu}{v (l_r + l_f)} C_{Sf} (g l_r - a h) \phi
\end{bmatrix}.
\]

The vehicle parameters are listed below:
\begin{itemize}
    \item \( \mu \): Surface friction coefficient: \(1.0489\)
    \item \( C_{Sf} \): Cornering stiffness coefficient, front: \( 4.718 /\, \text{rad} \)
    \item \( C_{Sr} \): Cornering stiffness coefficient, rear: \( 5.4562 /\, \text{rad} \)
    \item \( l_f \): Distance from center of gravity to front axle: \( 0.15875 \, \text{m} \)
    \item \( l_r \): Distance from center of gravity to rear axle: \( 0.17145 \, \text{m} \)
    \item \( h \): Height of center of gravity: \( 0.074 \, \text{m} \)
    \item \( m \): Total mass of the vehicle: \( 3.74 \, \text{kg} \)
    \item \( I \): Moment of inertia of the entire vehicle about the \( z \)-axis: \( 0.04712 \, \text{kg} \cdot \text{m}^2 \)
\end{itemize}